%% file: 00_main.tex
\documentclass[letterpaper]{article} % DO NOT CHANGE THIS
\usepackage{aaai2026}  % DO NOT CHANGE THIS
\usepackage{times}  % DO NOT CHANGE THIS
\usepackage{helvet}  % DO NOT CHANGE THIS
\usepackage{courier}  % DO NOT CHANGE THIS
\usepackage[hyphens]{url}  % DO NOT CHANGE THIS
\usepackage{graphicx} % DO NOT CHANGE THIS
\usepackage{natbib}  % DO NOT CHANGE THIS AND DO NOT ADD ANY OPTIONS TO IT
\usepackage{caption} % DO NOT CHANGE THIS AND DO NOT ADD ANY OPTIONS TO IT
\usepackage{algorithm}
\usepackage{algorithmic}
\usepackage{tabularx}
\usepackage{booktabs}
\usepackage{multirow}
\usepackage{array}
\newcolumntype{L}[1]{>{\raggedright\arraybackslash}p{#1}}
\usepackage{newfloat}
\usepackage{listings}
\DeclareCaptionStyle{ruled}{labelfont=normalfont,labelsep=colon,strut=off} % DO NOT CHANGE THIS
\floatstyle{ruled}
\newfloat{listing}{tb}{lst}{}
\floatname{listing}{Listing}
\title{Lost in Translation: How Universal Ethical Values Fail to Translate Across Global Contexts}
\author{
    Ozioma C. Oguine\textsuperscript{\rm 1},
    Munachimso B. Oguine\textsuperscript{\rm 2},
    Cesar Cervera\textsuperscript{\rm 1},
    Jenny Yang\textsuperscript{\rm 3},
    Pooja Voladoddi\textsuperscript{\rm 3},
    Mario Rodriguez\textsuperscript{\rm 3},
    Saif Eddin Bani Malhem\textsuperscript{\rm 3},
    Karla Badillo-Urquiola\textsuperscript{\rm 1},
    Daricia Wilkinson\textsuperscript{\rm 2}
}

\affiliations{
    \textsuperscript{\rm 1}University of Notre Dame\\
    \textsuperscript{\rm 2}Arizona State University\\
    \textsuperscript{\rm 3}Independent Researcher\\[0.5em]

    ooguine@nd.edu,
    moguine@asu.edu,
    ccervera@nd.edu,
    jennyhyang03@gmail.com,\\
    pooja.voladoddi@gmail.com,
    mariorodtcun@gmail.com,
    saif.banimalhem@insead.edu,\\
    kbadill3@nd.edu,
    daricia.wilkinson@asu.edu
}

\usepackage{bibentry}
\begin{document}

\maketitle

\begin{abstract}
%\textcolor{red}{DW: I'd be cautious with using principles, values, and ethics interchangeably throughout the text. The main focus was values as principles as more formal and generally perceived as less sensitive to cultural misinterpretations}- I agree, I have gone through the paper to adjust this

AI ethics frameworks treat values such as fairness, transparency, and accountability as universal and uniformly operationalizable across contexts. We examined how 14 experts across 10 countries made sense of AI in practice, reinterpreted core values, and envisioned governance alternatives. We found that AI deployment is characterized by structurally unequal conditions, marked by infrastructural constraints, extractive practices, and a ``mystification" of technology, which fundamentally shape perceptions of risks and opportunities. Our findings reveal that experts reinterpret values to fit local moral logics: privacy as collective and relational rather than individual; transparency as trust-building accountability rather than technical disclosure; and fairness as equity in access and representation rather than parity in outcomes. We identify these as translation gaps between encoded global frameworks and situated local practices. Finally, we propose pathways toward plural governance that redistributes epistemic authority and treats ethical negotiation as an ongoing, context-sensitive process rather than a settled technical standard.
\end{abstract}


\input{01_intro}
\input{02_background}
\input{03_method}
\input{04_findings}
\input{05_discussion}

\bigskip
% \noindent Thank you for reading these instructions carefully. We look forward to receiving your electronic files!

\bibliography{references}


\end{document}

%% file: 01_intro.tex
\section{Introduction}

As artificial intelligence (AI) systems are increasingly deployed across diverse global contexts \cite{statista2025}, questions of ethics and governance have become central to both research and policy \cite{aiindex2025, oecd2026, Oguine2026genai}. In response, a growing number of frameworks spanning institutions such as the OECD, UNESCO, and major technology companies have articulated core values including fairness, accountability, transparency, privacy, and safety \cite{oecd2026, unesco2021, hleg2019}. These frameworks are often presented as broadly applicable guidelines for responsible AI development and deployment. However, despite their widespread adoption, they are largely grounded in normative assumptions that may not translate seamlessly across cultural, social, and geopolitical contexts \cite{sambasivan2021, varshney2024}. Emerging scholarship has begun to critique the universalist framing of AI ethics, arguing that such approaches risk overlooking how values are interpreted, prioritized, and operationalized differently across communities \cite{mohamed2020, mhlambi2023}. In particular, perspectives from non-dominant regions, including the Global South and Indigenous communities, highlight how AI systems are entangled with historical inequities, cultural norms, and asymmetries in global power \cite{Nemorin2026decolonial_lens}. In these contexts, AI is not only a technical system but also a sociotechnical artifact that can reinforce or challenge existing structures of authority, representation, and autonomy \cite{mohamed2020}.

Despite these critiques, there remains limited empirical understanding of how experts \footnote{In this paper, We define \textbf{``experts"} in AI as individuals with both substantive engagement in AI (e.g., research, policy, industry, or advocacy) and deep familiarity with the cultural contexts in which AI is designed, deployed, or governed.} working across diverse cultural contexts, engage with dominant AI ethics frameworks in practice. Much of the existing literature has focused on defining ethical values or evaluating technical implementations \cite{jobin2019, morley2021initial}, with comparatively less attention to how these values are interpreted and adapted by those responsible for designing, governing, or critiquing AI systems in different regions. As a result, key questions remain about how AI ethics is interpreted across contexts and how local actors negotiate ethical values to align with their communities' needs. While existing critiques have focused on the abstraction of ethical principles or the challenges of implementing them in practice, considerably less attention has been paid to how these principles acquire meaning as they move across cultural, social, and institutional contexts. We argue that understanding AI ethics therefore requires attention not only to principles or implementation, but also to translation and interpretation in situated contexts. Specifically, we ask: \textbf{RQ1:} How do experts describe and make sense of AI deployment within their respective cultural contexts? \textbf{RQ2:} How do experts interpret and negotiate the ethical values embedded in AI systems? \textbf{RQ3:} How do they envision culturally grounded approaches to the design and governance of AI systems?

To answer these research questions, we conducted a qualitative interview study with 14 experts across 10 countries in Africa, Asia, Latin America, Oceania, Europe, North America, and the Caribbean. Our findings show that infrastructural constraints, competing narratives, and existing power dynamics shape how experts experience AI, perceived risks and opportunities. These experiences, in turn, inform how participants reinterpret core ethical values such as fairness, privacy, and transparency as relational, context-dependent, and embedded within broader social and structural conditions. Finally, participants identify key barriers in current governance approaches, alongside pathways that emphasize local knowledge, participatory processes, and more distributed forms of authority.

%\textcolor{red}{DW: values not principles i.e. make corrections in contribution statement} - DONE

This work makes three key contributions. This work makes three key contributions. First, we provide empirical evidence showing that the central challenge in AI ethics lies not only in defining universal values, but in how those values are interpreted across contexts. Second, we introduce the AI Ethics Translation Model, a conceptual framework that explains how universal ethical principles acquire situated meanings through local interpretation, producing translation gaps that shape governance outcomes. Third, we contribute to ongoing discussions in AI governance by identifying pathways toward more plural and power-aware approaches to AI governance, emphasizing the role of local knowledge, participatory processes, and more distributed forms of authority in shaping context-sensitive AI systems.

%% file: 02_background.tex
\section{Related Work}
The past decade has witnessed a proliferation of high-level AI ethics frameworks issued by international organizations and governmental bodies. The \citet{oecd2019}'s recommendation on Artificial Intelligence, endorsed by 47 countries, established foundational commitments to human-centered values, including transparency, fairness, and accountability. \citet{unesco2021} followed with its recommendation on the Ethics of AI, emphasizing human rights, dignity, and environmental sustainability as universal imperatives. The European Union's Ethics Guidelines for Trustworthy AI operationalized these principles through seven key requirements, including human agency, technical robustness, and diversity \cite{hleg2019}. These frameworks share the underlying assumption that ethical values for AI can be articulated at a sufficient level of abstraction to apply across diverse sociopolitical and cultural contexts.
However, scholars have increasingly critiqued this universalist orientation. \citet{greene2019} argue that such frameworks engage in ethical abstraction, distilling complex moral deliberation into decontextualized principles that fail to account for local power dynamics and value pluralism. Similarly, \citet{jobin2019}'s comprehensive survey of AI ethics guidelines reveals how fairness metrics predominantly encode Western liberal assumptions about individual autonomy and procedural justice, rendering them inadequate for contexts prioritizing communal harmony or relational ethics. The abstraction inherent in these frameworks, while enabling broad international consensus, risks producing surface-level compliance that masks substantive value misalignment, what \citet{Schultz2025Apr} calls ``ethics washing". These critiques suggest that the current architecture of global AI ethics may perpetuate epistemic injustice by privileging certain ways of knowing and valuing while marginalizing others.

Beyond critiques of universalism, researchers within the AIES community and beyond have begun articulating alternative epistemologies for AI ethics grounded in specific cultural traditions \cite{varshney2024, nicole2022ethics, schiff2020global-ethics, mhlambi2023}. For instance, research on African philosophies in computing has highlighted how Ubuntu ethics, which emphasize interconnectedness and communal flourishing, offer distinct frameworks for algorithmic fairness that diverge from individual rights-based approaches \cite{birhane2021, mhlambi2020, metz2022}. \citet{eke2022} recover forgotten African AI narratives that center communal values over individual autonomy, demonstrating the epistemic violence of excluding non-Western ethical traditions. Indigenous and First Nations scholars have also advanced particularly trenchant critiques of AI ethics, centering data sovereignty and epistemic self-determination. In their paper, \citet{lewis2020} articulated how Māori, Aboriginal, and Native American communities develop AI governance grounded in relational accountability to land, ancestors, and future generations. \citet{couldry2019} illuminates how the extraction of data from diverse cultural contexts mirrors historical resource extraction, undermining local value systems. 

The governance of AI systems remains concentrated among a relatively small set of institutions, shaping how ethical values are defined and applied across contexts. Prior work shows that systems and policies developed within specific settings often fail to account for diverse social norms, infrastructures, and knowledge systems \cite{hassan2023, dignazio2020}. Platform governance research further illustrates how these dynamics manifest in practice: content moderation policies encode particular assumptions about harm and expression \cite{gillespie2018}, while AI systems may overlook linguistic diversity, social hierarchies, and local usage practices \cite{sambasivan2021}. These mismatches reflect broader structural asymmetries in AI governance, raising questions about whose values are embedded in these systems and whose are excluded. Our study builds on these critiques by providing an empirical examination of how AI ethics is interpreted and operationalized across diverse cultural contexts. 

%% file: 03_method.tex
\section{Method} 
We employed a qualitative, interpretivist approach \cite{pervin2022interpretivist} to examine how experts across different cultural contexts interpret, negotiate, and envision culturally grounded approaches to AI ethics. This section outlines our participant recruitment, data collection procedures, analytical strategy, and ethical safeguards.

\subsection{Recruiting \& Participants}
We used purposive and convenience sampling to recruit experts with knowledge of both AI systems and their sociocultural contexts. Our sampling strategy aimed to capture perspectives across diverse global contexts. \textbf{Inclusion criteria} required participants to have: (1) expertise in AI development, deployment, or governance; and (2) deep familiarity with a specific cultural context, through lived experience or sustained professional engagement.

We recruited participants across multiple global regions, including Europe, North America, South America, Africa, Oceania, Asia, and the Caribbean, resulting in 14 experts from 10 countries. Recruitment was conducted through email outreach and social media platforms. We initially identified 62 potential participants who met our criteria; 17 agreed to participate, and 14 interviews were completed. The final sample (N=14) enabled in-depth exploration of cross-cultural perspectives while capturing variation across contexts. Table~\ref{tab:participants} summarizes participant demographics and regional distribution.

\begin{table*}[t]
\centering
\small
\begin{tabular}{L{0.7cm} L{1.9cm} L{3.5cm} L{3.7cm} L{3.8cm}}
\toprule
\textbf{ID} & \textbf{Country} & \textbf{Region} & \textbf{Sector} & \textbf{Area of Expertise} \\
\midrule
P1  & Mexico      & North/Central America & Industry                  & AI \\
P2  & Kenya       & Africa        & Tech Founder (Non-profit) & AI / Content Moderation \\
P3  & New Zealand & Oceania       & Industry                  & AI \\
P4  & USA     & North America        & Researcher \& Founder     & AI \\
P5  & New Zealand & Oceania       & Research Fellow           & AI \\
P6  & Zambia      & Africa        & Academia                  & Journalism / Emerging Media \\
P7  & Bangladesh  & South Asia    & Academia                  & HCAI \\
P8  & Ghana       & Africa        & Industry / Co-founder     & AI \\
P9  & Bangladesh  & South Asia    & Academia                  & HCI / AI \\
P10 & Nigeria     & Africa        & Researcher                & HCI \\
P11 & Nigeria     & Africa        & Researcher                & HCAI \\
P12 & France     & Europe     & Academia                  & AI Ethics \\
P13 & Jamaica     & Caribbean     & Academia                  & Humanities \\
P14 & Bangladesh  & South Asia    & Research Fellow           & HCAI \\
\bottomrule
\end{tabular}
\caption{Participant demographics across regions, sectors, and areas of expertise.}
\label{tab:participants}
\end{table*}

\subsection{Study Procedures \& Data Collection}
%\textcolor{red}{DW: add footnote with brief description and URL link to Dovetail e.g. Dovetail is an online tool for qualitative data analysis [insert url]}

We conducted semi-structured interviews with 14 experts over a seven-month period (December-June). Each interview lasted between 45-60 minutes and was conducted remotely via Zoom. With participant consent, interviews were audio-recorded, transcribed, and imported into Dovetail \footnote{Dovetail is an online tool for qualitative data analysis [https://dovetail.com/]} for qualitative analysis.
The interview protocol was designed to elicit culturally situated interpretations of AI ethics. To examine how experts engage with dominant AI ethics frameworks, we incorporated a structured probe centered on five widely cited values: bias, explainability, fairness, transparency, and privacy. These values were selected due to their prominence in existing AI ethics guidelines and policy documents identified in a prior systematic review we conducted (anonymized for review).
Participants were asked:
\begin{quote}
\textit{``Listed below are commonly cited AI ethics values. Please define each value based on your cultural context.''}
\end{quote}
Following this structured component, we used open-ended questions to explore additional culturally specific values and perspectives not captured by mainstream ethics frameworks. Probes focused on: (1) perceived risks and opportunities of AI development and deployment within participants’ cultural contexts; and (2) visions for culturally grounded approaches to designing and governing AI systems. The semi-structured format allowed flexibility to pursue emergent themes while ensuring consistent coverage of the research questions across interviews.

\subsection{Analysis Approach}
We employed a multi-stage qualitative analysis combining iterative coding with reflexive thematic analysis \cite{clarke2017thematic}.
\textbf{Stage 1: Initial Coding.} The first, second, third, and fourth authors independently reviewed all transcripts to achieve familiarity with the data. The fourth author then conducted open coding and annotation using Dovetail, generating an initial codebook that captured participants’ meanings and low-level concepts.
\textbf{Stage 2: Code Refinement and Categorization.} The first and second authors iteratively organized and refined initial codes to develop higher-level categories and identify relationships between concepts. This process involved constant comparison within and across transcripts to enhance conceptual clarity and coherence.
\textbf{Stage 3: Reflexive Thematic Analysis.} We then engaged in reflexive thematic analysis \cite{clarke2017thematic} to develop themes across the dataset. Here, we moved beyond descriptive coding to examine patterns of meaning, with particular attention to how cultural context shaped expert perspectives on AI ethics.
\textbf{Cross-Context Analysis.} To examine variation across contexts, we conducted both within-case and cross-case analyses \cite{moller2017explanatory}. Within-case analysis focused on the specificities of individual cultural contexts, while cross-case comparison identified convergent and divergent patterns across participants. This approach enabled us to preserve contextual nuance while developing broader analytical insights.
The analysis process was conducted through weekly meetings involving all authors over several months. These sessions supported reflexive discussion of coding decisions, theme development, and interpretive tensions. We maintained an audit trail documenting analytical decisions and revisions to the codebook to enhance transparency and rigor.

\subsection{Ethical Considerations}
This study was reviewed and approved by our institution’s ethics review board. Prior to any data collection, all participants signed a consent form that included agreement to audio record their session. At the start of each interview, we reconfirmed verbal consent and addressed any questions. We reminded participants that their engagement was entirely voluntary; they could pause, skip questions, or stop the session at any time and still receive the full thank-you gift. We protected all study data, including video recordings, audio files, notes, and transcripts, through multiple safeguards: encrypting all records at rest, restricting access to only the core research team and institutional administrators, and requiring two-factor authentication with a physical security key for data access. We retained only anonymized notes for use in the publication process. Finally, we asked each participant whether they would like to be recognized in acknowledgments or materials produced as part of the research. As a best practice, we attribute quotes only to participant IDs and specifically omit unique details, phrases, or words from quotes to mitigate identification of participants.

\subsection{Positionality Statement}
Our research team comprises authors from diverse cultural backgrounds across Africa, Asia, the Caribbean, the Middle East, South Asia, and Latin America. This diversity informed our interpretation of the data, particularly in how cultural context, power dynamics, and historical inequities shape perspectives on AI ethics. Several authors brought lived and professional experiences navigating marginalization in technology spaces, which sensitized our analysis to issues of representation, epistemic justice, and the uneven distribution of power in AI systems. We approached the analysis with an explicitly reflexive stance, recognizing that our positionalities both informed and shaped our interpretations. Throughout the coding and theme development process, we engaged in ongoing discussions to critically examine our assumptions, question interpretations, and account for how our cultural lenses influenced analytic decisions.

%% file: 04_findings.tex
\section{Findings}
We organize our findings to reflect how participants made sense of AI in their own contexts. We begin with how they described their experiences with AI in practice, particularly in terms of risks and opportunities (RQ1). These experiences then shaped how they interpreted key AI ethical values (RQ2). Finally, we examine how these insights informed their perspectives on culturally grounded approaches to AI governance (RQ3).

\subsection{Experts described AI Experience based on Risks and Opportunities (RQ1):}
Our analysis revealed that participants did not describe AI as a uniformly adopted or experienced technology. Instead, AI entered different contexts under uneven conditions shaped by infrastructural limitations, limited exposure, and uncertainty about how these systems function. In many cases, access to AI was contingent on broader technological ecosystems, where communities lacking stable electricity, connectivity, or devices were effectively excluded from meaningful engagement. As one participant noted:
\begin{quote}
    \textit{“They [tech corporations] would prefer to work with communities that have access to smartphones, electricity, [and] high-speed internet[...] but not every community is able to access some of this product.”} - P8
\end{quote}
This uneven entry into AI ecosystems is not simply a matter of availability, but reflects how development and deployment decisions are optimized for already-resourced environments. As a result, AI adoption is structured in ways that reproduce existing inequalities in access, shaping who benefits from these systems and who remains peripheral to them. Beyond infrastructure, participants emphasized that AI is often encountered through competing narratives of hype, fear, and uncertainty, which shape engagement even before adoption occurs. For example, one participant described how misinformation and sensationalized narratives led to fear-based interpretations of AI, noting that \textit{“there’s a lot of fear[...] teenagers were scared there’d be killer robots”} (P3). At the same time, others pointed to the opposite dynamic, where AI is framed in overly optimistic terms, making it \textit{“hard to think about the negative impacts”} (P4). These contrasting narratives shaped how participants made sense of AI in their contexts, with experiences of both uncertainty and optimism leading them to evaluate AI primarily in terms of its risks and opportunities. Across our analysis, four interrelated risk patterns emerged: (1) extractive data practices, (2) loss of local power and agency, (3) labor and economic exploitation, and (4) representation gaps and epistemic harm. Rather than functioning independently, these risks collectively reflect a broader dynamic in which AI systems extract value while limiting local control and representation.

At the same time, participants identified a set of opportunities that highlight AI’s potential when systems align with local needs and contexts. Four opportunity themes emerged: (1) expanding access to services and information, (2) supporting learning and knowledge acquisition, (3) improving efficiency in everyday tasks, and (4) enabling community empowerment and local innovation. We discuss these risks and opportunities in detail below.

\textbf{a) Risks in AI Deployment:}
Participants consistently described risks as emerging not only from the technical properties of AI systems, but also from how these systems interact with existing social, economic, and political structures. 
First, participants characterized \textbf{AI development as fundamentally extractive}, particularly in how data, cultural knowledge, and intellectual property are collected and repurposed without meaningful consent or governance. This extraction extends beyond raw data to include the contextual meanings embedded within that data, which are often stripped away during model training. As one participant explained, \textit{“AI is trained on data collected without people’s consent[...] they just pull data online[...] there is no transparency from development to deployment to training”} (P2). This highlights how AI systems rely on large-scale data aggregation practices that prioritize efficiency over accountability, raising concerns about ownership, consent, and the erasure of context.

Closely tied to this was a perceived \textbf{loss of local power and agency}. Participants described how AI systems are often designed externally and introduced into communities without meaningful opportunities for input or oversight, positioning local users as passive consumers rather than active contributors. As one participant mentioned:
\begin{quote}
    \textit{“Users don’t get to determine or be part of the designing process[...] we are more like[...] just use it.”} - P8
\end{quote}
Here, the concern is not only exclusion from design, but the broader reconfiguration of authority, where decision-making power is concentrated elsewhere. This reinforces asymmetries in who gets to define how AI systems operate and whose needs they prioritize.

Participants also emphasized \textbf{labor and economic exploitation} as a critical, yet often invisible, component of AI systems. Tasks such as data annotation and content moderation were described as underpaid, precarious, and psychologically harmful, particularly in contexts with limited labor protections. As one participant noted, \textit{“laborers are[...] exposed to violent and disturbing content[...] there’s absolutely no work protection[...] so that is also unfair”} (P9). These accounts reveal that AI systems are sustained by forms of labor that are both essential and systematically undervalued, raising questions about fairness not only in outputs, but in the conditions of production.

Finally, participants highlighted \textbf{representation gaps and epistemic harm} as a persistent risk, particularly in how AI systems fail to reflect local languages, accents, and cultural knowledge. These gaps were not framed solely as technical limitations, but as indicators of whose knowledge is prioritized in AI development. For instance, one participant explained that \textit{“we don’t have a chatbot that responds in Swahili[...] the accent[...] is not an African accent[...] so we feel we are left behind”} (P2). This implies that AI systems can marginalize users by failing to account for linguistic and cultural diversity, resulting in both practical exclusion and symbolic erasure. Overall, in addition to the new forms of harm AI might cause, participants also highlight its potential to amplify and reconfigure existing structural inequalities in data ownership, labor, power, and representation.

\textbf{b) Opportunities in AI Deployment:}
Despite these concerns, participants also described AI as offering meaningful opportunities, particularly when systems align with local needs, constraints, and priorities. Importantly, these opportunities were not framed as universal or inherent to AI, but as context-dependent outcomes that emerge under specific conditions of alignment between technology and use context.

Participants highlighted the role of \textbf{AI in supporting learning and education}, particularly as a tool for augmenting rather than replacing existing practices. AI was seen as enabling new forms of engagement when integrated thoughtfully. As one participant noted, AI \textit{“is to enhance learning[...] not to duplicate it[...] but to support what’s already there”} (P13). This reflects a broader orientation toward AI as a complementary resource, rather than a disruptive force.

In addition, AI was described as \textbf{expanding access to essential services}, particularly in contexts where traditional infrastructure is limited. Participants pointed to examples such as mobile banking and e-governance systems as ways in which AI-enabled technologies can extend reach. One participant explained:
\begin{quote}
    \textit{“The phone itself is a bank[...] you don’t need to have a smartphone[...] or even a bank.”} - P6
\end{quote}
This illustrates how AI can be embedded within existing infrastructures to lower barriers to access, particularly in resource-constrained environments.

Participants also emphasized AI’s role in improving efficiency in everyday tasks, where the value of AI is often tied to convenience rather than technical understanding. As one participant noted, \textit{“they don’t really care how the system is working[...] it’s convenient for them, and that’s everything they care about”} (P7). This highlights how adoption is often driven by practical utility, even in the absence of deep technical knowledge.

Finally, participants described AI as a potential tool for community empowerment, particularly when communities are able to shape how systems are developed and used. In such cases, AI can serve as a mechanism to amplify local voices and address existing biases. As one participant suggested, \textit{“if we can work with the data[...] we can empower our culture and our people”} (P3). This points to the possibility of AI not only reinforcing inequalities, but also supporting more equitable forms of participation and representation when aligned with local control. Overall, these opportunities demonstrate that AI’s benefits are not guaranteed, but emerge when systems are designed and deployed in ways that reflect the specific needs and contexts of the communities they serve.

%===========================================================================================================================================================

\begin{figure*}[t]
    \centering
    \includegraphics[width=0.65\textwidth]{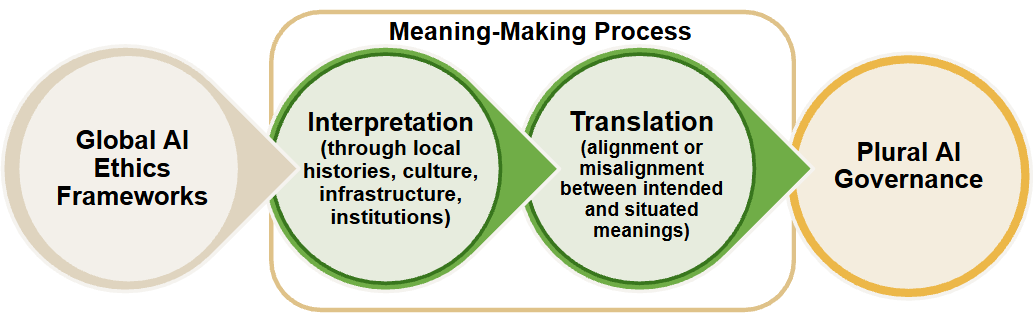}
    \caption{\textbf{Conceptual model of AI ethics translation.} Universal AI ethics frameworks acquire context-specific meanings through local interpretation. This process may produce alignment or translation gaps, which subsequently shape culturally grounded AI governance.}
    \label{fig:translation-model}
\end{figure*}

\subsection{Ethical Values Interpreted Differently Across Cultural Contexts (RQ2)}
Building on participants’ experiences with AI in practice, we found that interpretations of core AI ethical values varied significantly across contexts. Participants reflected on widely cited values such as bias, privacy, fairness, transparency, and explicability not as fixed or universal values, but as concepts whose meanings are shaped by cultural norms, lived realities, and structural conditions. Across interviews, participants challenged dominant assumptions embedded in existing AI ethics frameworks, particularly those grounded in Western, individualistic, and decontextualized understandings of these values. Instead, they articulated interpretations that emphasized relational, contextual, and collective dimensions of ethical practice. In doing so, participants also introduced locally grounded values such as community, respect, and collective well-being, which they viewed as essential yet often overlooked in mainstream AI ethics discourse.

Our analysis identified five major themes in how participants interpreted AI ethical values: (1) reinforcement of marginalized and colonial ideals, (2) privacy as collective and relational practice, (3) transparency as a foundation for trust and accountability, (4) fairness as equity and structural justice, and (5) explicability as meaningful and contextual interpretation.

\textbf{a) Reinforcement of Marginalized and Colonial Ideals:}
Most participants interpreted bias not simply as a technical issue of data imbalance, but as a form of ``\textit{epistemic imperialism}" that reflects the broader power dynamics embedded within AI systems. Rather than viewing bias as an isolated flaw, participants emphasized how AI systems reproduce and reinforce existing social hierarchies by prioritizing certain knowledge systems, values, and identities over others. In this sense, bias was understood as structural, emerging from historical patterns of exclusion that are encoded into data and algorithms. For instance, P9 noted, \textit{"when you try to apply your understanding of ethics in a culture, which you have no idea about. You are basically doing like practicing colonization, then a colonial practice, you are telling them what is right and what is wrong"}.

Participants emphasized that many AI systems are trained on datasets that fail to adequately represent local contexts, leading to outputs that are misaligned with users’ realities. As one participant noted, models are often “\textit{trained on data that does not represent the community}” (P8), resulting in systems that fail to produce relevant or accurate outcomes. This lack of representation was not seen as incidental, but as indicative of deeper imbalances in whose data is collected and whose knowledge is valued. Beyond representation, participants described how AI systems embed and enforce dominant cultural norms, particularly those associated with Western perspectives. One participant explained:
\begin{quote}
    \textit{“If it aligns with Western values, then it’s good. If it does not align[...] then it’s bad[...] we have to tell you what is human rights.”} - P6
\end{quote}
Here, bias is not only about misrepresentation, but about the imposition of particular moral frameworks as universal standards. Participants framed this as a continuation of colonial dynamics, where certain epistemologies are privileged while others are marginalized. Together, these accounts demonstrate that bias in AI systems is not merely technical but deeply tied to questions of power, representation, and epistemic authority.

\textbf{b) Privacy as Collective and Relational Practice:}
Participants challenged dominant interpretations of privacy as an individual right, instead framing it as a collective and relational practice shaped by social norms, family structures, and community expectations. In many contexts, privacy was not understood as control over personal data in isolation, but as something negotiated within relationships and shared social spaces. As one participant explained, assumptions that marginalized communities do not value privacy are misleading, noting that:
\begin{quote}
    \textit{“So in the West, we would say the individual is everything, right? It's about my privacy. Whereas in eastern and indigenous cultures, it's much more of a group dynamic. And so you have extended family unit in Maori culture. Um, we talk about us, rather than I.”} - P9
\end{quote}
This highlights a disconnect between how privacy is operationalized in AI systems, typically as individual data ownership, and how it is experienced in practice. Participants emphasized that privacy decisions are often made collectively, taking into account the well-being and reputation of families or communities. Additionally, privacy was closely tied to trust and social accountability. Participants noted that sharing data is often contingent on relationships and familiarity, rather than formal consent mechanisms.

\textbf{c) Transparency as a Foundation for Trust and Accountability:}
Participants interpreted transparency not simply as access to information about how AI systems operate, but as a foundation for building trust and ensuring accountability. Rather than focusing on technical explanations alone, participants emphasized the importance of being able to understand, question, and trace decisions made by AI systems. A recurring concern was that many AI systems are perceived as opaque and difficult to engage with, which undermines trust. As one participant noted, \textit{“you can’t see it, you can’t touch it[...] I’m not trusting this thing”} (P13). This highlights how abstract and intangible systems challenge expectations of accountability, particularly in contexts where trust is built through direct interaction and visibility. Participants also emphasized that explanations must be meaningful within local contexts. Technical descriptions were often insufficient, particularly when they did not align with users’ knowledge or experiences. One participant illustrated this through the example of agricultural AI tools:

\begin{quote}
    \textit{“When farmers ask how [AI] knows whether it’s going to rain[...] no one can explain it[...] so they become uncomfortable sharing their data.”} - P9
\end{quote}
Participants viewed transparency as a mechanism for enabling trust, where users can make informed decisions about whether to engage with AI systems.

\textbf{d) Fairness as Equity and Structural Justice:}
Participants interpreted fairness not as equality of outcomes, but as equity in inclusion, emphasizing the need to account for structural inequalities in how AI systems are designed and deployed. Rather than treating fairness as a neutral or technical property, participants framed it as inherently political, shaped by access to resources, representation in data, and participation in decision-making.
Participants highlighted that AI systems designed without considering infrastructural disparities may be technically fair but practically exclusionary. For example, access to AI services often depends on connectivity, literacy, and economic resources, raising questions about who benefits from these systems. As one participant noted, \textit{“you can’t call it fair if people cannot even access it,”} emphasizing that fairness must be evaluated in relation to real-world conditions.

Furthermore, fairness was linked to representation and inclusion in system design. Participants stressed the importance of ensuring that diverse voices are involved in shaping AI systems, rather than relying solely on external perspectives. This reflects a broader understanding of fairness as a process, rather than an outcome, one that requires ongoing attention to equity, participation, and structural conditions.

\textbf{e) Explicability as Meaningful and Contextual Interpretation:}
Participants emphasized that explicability extends beyond technical transparency to include the ability to communicate AI processes in ways that are culturally meaningful and accessible. While many systems provide outputs with high confidence, participants noted that users often lack the ability to assess their validity or understand how they were generated. As one participant explained:

\begin{quote}
    \textit{“The system gives very confident answers[...], but you have no idea what went into that[...] the only way to know is if you already knew the answer.”} - P5
\end{quote}

This highlights a gap between system outputs and user understanding, where explanations fail to support meaningful engagement. Participants emphasized that effective explicability requires grounding explanations in local languages, cultural references, and everyday experiences. Rather than expecting users to adapt to technical systems, participants argued that AI should adapt to users’ ways of knowing. This includes designing explanations that are intuitive, contextually relevant, and aligned with users’ interpretive frameworks.

In addition to reinterpreting mainstream AI ethics values, participants also \textbf{articulated locally grounded values} that they viewed as essential to ethical AI design and deployment. These values, such as respect, community, and collective well-being, extend beyond dominant frameworks that prioritize individual rights and technical performance. Instead, they emphasize relational accountability, social cohesion, and alignment with cultural norms. Participants described these values as shaping how technologies should function within their communities, particularly in ensuring that systems are not only technically effective but also socially appropriate and respectful of local practices. As one participant noted:
\begin{quote}
    \textit{“There are also cultural practices that AI systems fail to capture or respect, such as women covering their faces for religious or cultural reasons”} - P9
\end{quote}
This implies that ethical AI cannot be fully captured through universal values, but must also account for locally embedded value systems and meaning-making that define what is considered acceptable, respectful, and beneficial in specific contexts. Participants consistently described how global AI ethics frameworks were interpreted through local histories, cultural norms, infrastructures, and institutional realities. We define \emph{translation gaps} as the divergence between the intended meanings of universal AI ethical principles and their situated interpretations across contexts. Figure \ref{fig:translation-model} synthesizes this conceptual process, illustrating how universal AI ethics frameworks are interpreted through local cultural, historical, infrastructural, and institutional contexts, and how these situated interpretations inform more plural and culturally grounded approaches to AI governance.

%==========================================================================================================================================================

\subsection{Barriers and Pathways to Culturally and Plurally Grounded AI Governance (RQ3)}
Drawing on participants’ experiences with AI in practice (RQ1) and their interpretations of core ethical values (RQ2) as depicted in Figure \ref{fig:translation-model}, participants framed AI governance as a sociotechnical challenge shaped by misalignment between existing frameworks and local realities. Rather than viewing governance as a matter of implementing predefined values, participants emphasized the need to reconsider how ethical values are defined, operationalized, and enforced across contexts. Across our analysis, participants identified key barriers embedded within current approaches to AI ethics as well as pathways for rethinking governance in ways that are more culturally grounded and context-sensitive.

\textbf{a) Barriers to Culturally Grounded AI Governance:}
Participants described three primary barriers that constrain the development of culturally grounded AI governance: (1) misalignment between ethical frameworks and contextual value systems, (2) institutional structures that limit adaptation, and (3) structural inequalities that shape participation.

Many participants emphasized that widely adopted AI ethics frameworks often \textbf{fail to translate across contexts} because they are built on specific cultural assumptions about values such as harm, fairness, and appropriateness. These frameworks tend to operationalize ethics through fixed categories, while participants’ interpretations of these values are relational, contextual, and dynamic (RQ2). As a result, governance systems enforce standards that may conflict with local norms and practices. For example, one participant described how content moderation policies reflect external interpretations of acceptable behavior:
\begin{quote}
    \textit{“How we perceive nudity is different… even our laws are different[...] they bring their own rules.”} - P8
\end{quote}
This highlights how governance frameworks can impose singular interpretations of ethical values, overriding local legal and cultural systems. Similarly, linguistic expressions are often interpreted out of context, where socially embedded forms of communication are misclassified as harmful. These tensions reveal that governance challenges emerge not from a lack of values, but from how those values are translated into system rules and policies.

Next, participants further described how \textbf{governance is shaped by institutional systems that codify narrow interpretations} of ethical values, limiting flexibility in how these values can be applied. Legal and regulatory frameworks often reflect dominant epistemologies, making it difficult to incorporate alternative understandings. One participant illustrated this through privacy regulations:
\begin{quote}
    \textit{“Privacy laws[...] only recognize the individual[...] but from our perspective, privacy includes the collective.”} - P9
\end{quote}
Additionally, participants highlighted how \textbf{governance decisions are influenced by centralized sources of authority}, particularly industry and external policy actors. As one participant explained, decision-makers are often exposed primarily to optimistic narratives about AI, which shape regulatory priorities and limit critical engagement. Together, these dynamics reinforce governance models that are difficult to adapt to diverse contexts.

Lastly, participants emphasized that participation in AI governance was shaped by broader structural conditions, including access to resources, knowledge, and infrastructure. Engagement with governance processes requires a baseline level of awareness and capacity, which is unevenly distributed across contexts. As one participant explained, \textit{“you can’t really talk about trust unless people understand what AI is”} (P13), highlighting how limited familiarity with AI constrains meaningful participation. In addition, competing priorities, such as meeting basic needs, reduce the feasibility of engaging with governance discussions. These conditions shape whose voices are represented in governance processes, reinforcing existing inequalities in decision-making.

\textbf{b) Pathways Toward Culturally and Plurally Grounded AI Governance.}
In response to these barriers, participants articulated pathways that move beyond adapting existing frameworks toward reconfiguring how AI governance is structured. Four key themes emerged: (1) Centering Local Knowledge in AI Design and Evaluation, (2) Enabling Co-Constructed Governance Through Participatory Approaches, (3) Community Sovereignty over Data, Knowledge, and Representation, and (4) Rebalancing AI Knowledge and Governance Across Contexts.

Participants expressed the need and urgency of \textbf{grounding AI systems in local epistemologies}, rather than adapting externally developed models to local contexts. This involves recognizing local knowledge as a foundation for design, rather than as supplementary input. For instance, one participant described how existing cultural frameworks could guide AI development:
\begin{quote}
    \textit{“We should be creating frameworks for AI development[...] much like our own research methods.”} - P14
\end{quote}
This reflects a shift toward designing systems that emerge from local ways of knowing, enabling more contextually aligned interpretations of ethical values.

Several participants also highlighted the \textbf{need for participatory governance models} that involve communities in shaping both system design and policy. Participation was framed as co-creation rather than consultation, where stakeholders actively contribute to defining ethical priorities. As one participant explained, \textit{“you need to work with communities[...] ethics is not universal”} (P3), emphasizing that ethical values must be negotiated rather than imposed. This approach enables governance systems to reflect diverse perspectives and adapt to local contexts.

Participants further emphasized the importance of \textbf{community control over data and knowledge}, framing it as essential for addressing extractive practices identified in RQ1. Data sovereignty was described as a mechanism for ensuring that communities retain authority over how their data is used. One participant noted that \textit{“ensuring consent and control over data[...] would be a good start”} (P2), highlighting how governance must extend beyond access to include ownership and accountability. This reframes data governance as a question of stewardship and responsibility.

Finally, participants described the need to \textbf{rebalance global AI governance by redistributing authority} and recognizing multiple knowledge systems. This involves moving away from centralized models toward more distributed and context-sensitive approaches. As one participant explained:
\begin{quote}
    \textit{“We need to stop extractive practices[...] build locally[...] and learn from experts within those contexts.” }- P6
\end{quote}
This reflects a broader shift toward plural governance, where ethical values are not standardized but negotiated across contexts. Overall, participants framed AI governance as an ongoing process of negotiating across diverse value systems, rather than implementing universal principles. Addressing the identified barriers and advancing these pathways requires rethinking how ethical values are translated, how authority is distributed, and how participation is structured in AI systems. Synthesizing our findings across RQ1-RQ3, we present a conceptual overview of how participants’ experiences with AI shape the interpretation of ethical values, where misalignments emerge, and how these tensions inform governance implications. This synthesis highlights key patterns that connect experiences, interpretations, and governance across contexts.

%% file: 05_discussion.tex
\section{Discussion}
In this section, we situate our findings within broader conversations in AI ethics and governance, focusing on how participants’ experiences, interpretations of ethical values, and proposed pathways reveal key limitations in current approaches. We highlight two key contributions that both align with and extend prior work.

\subsection{Rethinking deficit framing in AI ethics positions local knowledge as Governance Infrastructure}
In our findings, participants did not position themselves or their communities as lacking ethical understanding of AI. Instead, they articulated rich, contextually grounded interpretations of values, alongside locally rooted values such as community, respect, and collective well-being. These perspectives were not framed as alternatives to mainstream AI ethics, but as already-existing systems of meaning that shape how technology is evaluated and used in practice.  Dominant trends in AI ethics has focused on identifying and aggregating shared ethical principles such as fairness, transparency, and accountability across institutions and national contexts \cite{Hagendorff2022Nov, jobin2019}. These efforts have been instrumental in establishing a common vocabulary for AI governance, and in this sense, our findings align with these prior works in recognizing the continued relevance of these values. 

However, many of the prior papers also rely on an implicit assumption that once defined, these values can be uniformly interpreted and operationalized across contexts \cite{Hagendorff2022Nov}. Our findings complicate this assumption. Participants consistently demonstrated that the central challenge is not defining ethical values, but translating them into practice in ways that align with local contexts. This resonates with longstanding scholarly critiques of abstraction, particularly \cite{Tidjon2022May, Goffi2022cultural-ethics}, which argue that systems often fail when they impose simplified representations onto complex social realities. In our study, this abstraction manifests as what we conceptualize as \emph{\textbf{``translation gaps"}}: which are divergences between the meanings intended by universal AI ethics frameworks and the situated interpretations these values acquire across cultural, historical, and institutional contexts.

Crucially, these gaps are not only interpretive but material. As seen in RQ1 and RQ3, they shape how systems are adopted, trusted, and resisted. Misalignment leads to misclassification of culturally situated behaviors, breakdowns in trust, and governance systems that feel externally imposed rather than locally relevant. This extends prior critiques of “ethical abstraction” \cite{greene2019, Munn2023Aug} by showing how these issues are experienced in practice across diverse contexts.
At the same time, our findings challenge the prevailing “inclusion” paradigm in HCI and AI ethics \cite{Dine2025inclusion, oguine2025inclusion}. Much prior work \cite{mayeesha2025bangladesh, nicole2022ethics} has framed marginalized communities as needing access, representation, or inclusion within existing systems (e.g., digital divide literature, participatory design). While these efforts are important, our findings suggest that inclusion alone is insufficient. In many cases, inclusion operates within systems whose underlying assumptions remain unchanged, resulting in what participants described as forms of epistemic extraction where local data, knowledge, and perspectives are incorporated without shifting who defines the system \cite{Munn2023Aug}. Our findings suggest that when communities are included only as data sources or consulted without veto power, AI ethics risks becoming a form of 'participatory extraction' in which local nuances are harvested to make global systems more resilient, without ceding any structural authority to local actors. Hence, participants’ interpretations of values and their articulation of locally grounded values demonstrate that governance can be built from these knowledge systems, rather than layered onto them. This shifts the focus from “how do we include diverse users?” to “how do we design governance systems that emerge from diverse epistemologies?”

\subsection{Moving from Universalism to Plural, Power-Aware AI Governance}
Our findings also speak directly to ongoing debates about universalism in AI ethics \cite{Floridi2019unified, Munn2023Aug}. Many global frameworks, like the OECD and UNESCO, operate on the premise that ethical alignment can be achieved through convergence on shared values \cite{Munn2023Aug, birhane2022forgotten}. This model of “value universalism” assumes that concepts such as fairness or transparency possess stable meanings that can be applied across contexts \cite{Floridi2019unified, mohamed2020}. Participants in our study did not reject these values. Instead, they revealed how these values take on multiple, co-existing meanings depending on context. As illustrated in Figure\ref{fig:translation-model}, governance emerges only after ethical values have been interpreted within local contexts. This suggests that supporting plural AI governance requires designing mechanisms that acknowledge and negotiate these situated interpretations, rather than assuming that shared principles will be uniformly understood. These interpretations align with growing critiques in HCI and AI ethics that question the universality of ethical values \cite{greene2019, Khan2021Sep}, while providing empirical grounding for how these tensions manifest in practice.

Importantly, our findings suggest that the challenge of AI governance is not simply cultural, but deeply tied to power and epistemic authority. Decisions about how ethical values are defined, operationalized, and enforced remain concentrated within specific institutions, regions, and industries. This reflects broader patterns of epistemic inequality, where certain forms of knowledge are privileged while others are marginalized \cite{birhane2022forgotten, dignazio2020, Munn2023Aug}. As participants described, local actors are often positioned as adopters of externally developed systems rather than contributors to their design and governance. This insight extends existing calls for participatory and inclusive AI governance by emphasizing that participation alone does not address underlying power asymmetries. Instead, what is needed is a shift toward plural and power-aware governance, where multiple value systems are not only represented but actively shape decision-making processes. In this model, governance is not about enforcing consensus on a single interpretation of ethical values, but about enabling coordination across diverse perspectives.
The pathways identified in our findings, such as participatory governance, local knowledge integration, and data sovereignty, point toward how this shift might be realized in practice. These approaches redistribute authority, recognize diverse forms of expertise, and create space for ongoing negotiation of values. Rather than treating ethical values as fixed standards to be implemented, plural AI governance requires institutional mechanisms that support the ongoing interpretation, negotiation, and revision of ethical values across contexts.

\subsection{Implications for AI Design, Policy and Governance}
We outline implications that follow from our findings, focusing on how AI systems, governance frameworks, and participation models must shift to better support the contextual interpretation and plural negotiation of ethical values.

\textbf{Design Implications:} Our findings suggest that AI system design must move beyond treating ethical values as fixed inputs and instead support their contextual interpretation in use and systemic contestability \cite{Alfrink2023contestability}. Prior work within AIES and allied fields \cite{varshney2024, delgado2023stakeholder, whittlestone2019ethical} has emphasized the importance of designing for situated action and meaning-making, yet many AI systems continue to rely on static representations of harm, fairness, and appropriateness. Designing for translation requires developing systems that can accommodate multiple interpretations of ethical values, potentially through pluralistic auditing, localized models, or mechanisms that allow users to contest and reinterpret system outputs through local community councils or groups. This shifts the role of AI from enforcing predefined norms to supporting ongoing negotiation of meaning, which is critical in culturally diverse and globally deployed systems.

\textbf{Governance and Policy Implications:} At the policy level, our findings challenge the dominance of one-size-fits-all governance frameworks. While global guidelines provide useful high-level principles, they often lack mechanisms for contextual adaptation. Subsidiarity in AI governance should therefore operate across multiple levels, allowing ethical values to be interpreted and operationalized locally while maintaining broader coordination \cite{capp2026subsidiarity}. This aligns with emerging discussions in AI policy that advocate for flexible and context-sensitive regulatory approaches \cite{whittlestone2019ethical}. Importantly, such models must also account for the structural conditions identified in RQ2, ensuring that governance frameworks are not only adaptable in principle but feasible in practice across different infrastructural and socio-political contexts.

\textbf{Implications for Participation and Stakeholder Engagement:}
Our findings also reinforce the need to move from consultation-based approaches toward co-constructive models of participation. While participatory design has long been a focus in HCI, its application in AI governance remains limited and often superficial \cite{Oguine2026genai, sloane2022participation}. Participants in this study emphasized that meaningful engagement requires not only inclusion but also the ability to influence decision-making processes. This requires addressing structural barriers such as access, literacy, and resource constraints that limit participation, as well as creating sustained mechanisms for engagement rather than one-off consultations. In doing so, governance becomes a continuous, collaborative process that evolves with community needs and interpretations.

\textbf{Implications for Data Practices and Governance:} The findings further highlight the need to reconceptualize data as socially and culturally embedded, rather than as neutral inputs to AI systems \cite{Loukissas2019data}. Existing data practices often prioritize extraction and aggregation, overlooking the relationships and meanings associated with data in different contexts. Participants’ emphasis on data sovereignty aligns with growing calls for more equitable and accountable data governance models \cite{mhlambi2020, kukutai2016indigenous}. Supporting community control over data requires not only technical solutions but also institutional changes that enable ownership, consent, and stewardship at the community level. This shifts data governance from a model of access to one of accountability and reciprocity.

\section{Limitations and Future Work}
This work should be interpreted in light of several limitations that also point to future research. First, our purposive sample of 14 experts enabled the depth of qualitative analysis we sought. While the findings are not intended to be generalizable, they offer theoretically transferable insights into how ethical translation operates across diverse contexts. Accordingly, the perspectives presented here should be understood as situated interpretations shaped by participants' professional and lived experiences, rather than representative accounts of the cultures in which they are situated. Future work should examine how translation operates across a broader range of stakeholders, particularly marginalized and non-expert communities. Second, our findings rely on self-reported accounts, which may differ from how AI ethics is negotiated in practice. Ethnographic or longitudinal studies could examine these dynamics over time. Finally, while this paper highlights the need for plural AI governance, it remains primarily analytical. Future work should explore governance mechanisms that balance locally situated interpretations of ethical values with globally shared commitments to human rights, safety, and accountability.

\section{Conclusion}
This paper examined how experts across diverse contexts experience AI systems, interpret core ethical values, and envision pathways toward more culturally grounded governance. Our findings show that widely cited AI ethics principles, such as fairness, are not universally interpreted, but are shaped by contextual and relational factors. We argue that the central challenge in AI ethics is not defining shared values, but translating them across contexts. These translation gaps contribute to misalignment between systems and the communities they serve, reinforcing the need for more context-sensitive approaches. In response, participants highlighted pathways that emphasize local knowledge, participatory processes, and more distributed forms of governance. Hence, this work reframes AI governance as an ongoing process of negotiating ethical values across diverse contexts, pointing toward the need for more plural and power-aware approaches to designing and governing AI systems.

\section*{Acknowledgments}
We sincerely thank all of our participants for sharing their time, experiences, and perspectives for this study. Their thoughtful reflections made this work possible.